\documentclass[letterpaper]{article} 
\usepackage{aaai24}  
\usepackage{times}  
\usepackage{helvet}  
\usepackage{courier}  
\usepackage[hyphens]{url}  
\usepackage{graphicx} 
\usepackage{natbib}  
\usepackage{caption} 
\usepackage{algorithm}
\usepackage{booktabs}
\usepackage{multirow}
\usepackage{multicol}
\usepackage{makecell}
\usepackage{color}
\usepackage{colortbl}
\usepackage{soul}
\definecolor{lightred}{rgb}{1, 0.8, 0.8}
\definecolor{lightblue}{rgb}{0.8, 0.9, 1}

\newcommand*\colourcheck[1]{%
  \expandafter\newcommand\csname #1check\endcsname{\textcolor{#1}{\ding{52}}}%
}
\colourcheck{blue}
\colourcheck{green}
\colourcheck{red}
\usepackage{pifont}
\newcommand{\xmark}{\ding{55}}%
\usepackage{amsmath}
\usepackage{graphicx}
\usepackage{amsfonts}
\usepackage{algpseudocode}
\DeclareMathOperator*{\argmax}{arg\,max}

\usepackage{placeins}

\usepackage[most]{tcolorbox}
\usepackage{listings}
\usepackage{float}

\tcbset{
  aibox/.style={
    width=474.18663pt,
    top=10pt,
    colback=white,
    colframe=black,
    colbacktitle=black,
    enhanced,
    center,
    attach boxed title to top left={yshift=-0.1in,xshift=0.15in},
    boxed title style={boxrule=0pt,colframe=white,},
  }
}
\newtcolorbox{AIbox}[2][]{aibox,title=#2,#1}

\usepackage{newfloat}
\usepackage{listings}
\DeclareCaptionStyle{ruled}{labelfont=normalfont,labelsep=colon,strut=off} 
\floatstyle{ruled}
\newfloat{listing}{tb}{lst}{}
\floatname{listing}{Listing}
\title{Improving Scientific Hypothesis Generation with Knowledge Grounded Large Language Models}

\author{
    Guangzhi Xiong\textsuperscript{\rm 1}\equalcontrib, Eric Xie\textsuperscript{\rm 1}\equalcontrib, Amir Hassan Shariatmadari\textsuperscript{\rm 1}, Sikun Guo\textsuperscript{\rm 1}, \\Stefan Bekiranov\textsuperscript{\rm 1}, Aidong Zhang\textsuperscript{\rm 1}
}
\affiliations{
    \textsuperscript{\rm 1}University of Virginia
}

\usepackage{bibentry}

 \usepackage{amsmath}
\usepackage{xcolor}

\begin{document}

\maketitle

\begin{abstract}
Large language models (LLMs) have demonstrated remarkable capabilities in various scientific domains, from natural language processing to complex problem-solving tasks. Their ability to understand and generate human-like text has opened up new possibilities for advancing scientific research, enabling tasks such as data analysis, literature review, and even experimental design. One of the most promising applications of LLMs in this context is hypothesis generation, where they can identify novel research directions by analyzing existing knowledge. However, despite their potential, LLMs are prone to generating ``hallucinations'', outputs that are plausible-sounding but factually incorrect. Such a problem presents significant challenges in scientific fields that demand rigorous accuracy and verifiability, potentially leading to erroneous or misleading conclusions. To overcome these challenges, we propose KG-CoI (Knowledge Grounded Chain of Ideas), a novel system that enhances LLM hypothesis generation by integrating external, structured knowledge from knowledge graphs (KGs). KG-CoI guides LLMs through a structured reasoning process, organizing their output as a chain of ideas (CoI), and includes a KG-supported module for the detection of hallucinations. With experiments on our newly constructed hypothesis generation dataset, we demonstrate that KG-CoI not only improves the accuracy of LLM-generated hypotheses but also reduces the hallucination in their reasoning chains, highlighting its effectiveness in advancing real-world scientific research.
\end{abstract}

\vspace{-0.07in}
\section{Introduction}
Advanced Large Language Models (LLMs) such as GPT-4 \cite{openai2024gpt4technicalreport} have exhibited exceptional performance in a wide range of general machine learning tasks, such as question answering \cite{hendrycks2021measuring} and arithmetic computation \cite{cobbe2021training}. Recently, there has been growing interest in harnessing the reasoning capabilities of LLMs within scientific domains. 
These efforts have shown impressive results, with LLMs tackling complex scientific questions and often achieving, or even exceeding, human-level performance \cite{nori2023capabilities, hou2023assessing, stribling2024model}.
As a result, LLMs are increasingly viewed as promising tools with the potential to significantly advance real-world scientific research, such as drug discovery, biological sequence analysis, and material design \cite{ai4science2023impact}.

Specifically, with the capability of processing and synthesizing vast amounts of text, LLMs are well-suited to accelerate the analysis of scientific literature and generate new hypotheses for potential scientific discovery \cite{qi2023large,zhou2024hypothesis}. Traditional scientific research, particularly in natural sciences like biology, typically involves a multi-step, time-consuming process from gathering literature to validating hypotheses. By generating promising research ideas directly from existing literature, LLMs have the potential to significantly streamline and reduce the time required for these labor-intensive tasks.

However, despite their advanced capabilities, LLMs face criticism for generating misinformation or so-called ``hallucinations'', which are responses that seem plausible but are factually incorrect \cite{huang2023survey}. This issue is particularly critical in scientific research, where every reasoning step must be transparent and verifiable. In the context of hypothesis generation for natural sciences, hallucinations can easily arise if the parametric knowledge of LLMs lacks accurate scientific information. Moreover, these hallucinations are particularly challenging to detect in generated hypotheses, as they are often related to potential discoveries that have not yet been explored. 

To address the problems mentioned above, we propose a novel system termed KG-CoI (Knowledge Grounded Chain of Ideas), designed to enhance hypothesis generation by incorporating external knowledge from knowledge graphs (KGs). These KGs contain well-organized structured information that has been verified by existing literature. By prompting LLMs to generate a chain of ideas (CoI) through step-by-step reasoning \cite{wei2022chain}, our system facilitates in-depth analysis of the input and further verification of the generated content. The KG-CoI system consists of three key modules: KG-guided context retrieval, KG-augmented chain-of-idea generation, and KG-supported hallucination detection. By linking LLM hypothesis generation to KGs, our system aligns the output with well-established scientific knowledge and ensures that the generated hypotheses are grounded in reliable information sources. 

To quantitatively demonstrate the effectiveness of our system, we construct a hypothesis generation dataset by masking certain links within a KG and prompting LLMs to hypothesize potential relations without prior knowledge of the facts. Our experiments show that, compared to existing methods for prompting LLMs in hypothesis generation, KG-CoI achieves the highest accuracy in generating hypotheses, underscoring its advantages in real-world scientific research. Moreover, with the KG-supported hallucination detection, we demonstrate the effectiveness of KG-CoI in reducing hallucinations, thereby improving its reliability in natural sciences.

Our contributions can be summarized as follows:
\begin{itemize}
\item We present KG-CoI, a novel LLM-enhanced hypothesis generation system that augments the generated hypotheses with external structured knowledge and presents the result as a coherent chain of ideas.
\item We construct a new dataset to evaluate LLM hypothesis generation and conduct extensive experiments on both open- and close-source LLMs, showing the effectiveness of KG-CoI in hypothesizing scientific knowledge.
\item We propose a KG-supported hallucination detection method within KG-CoI, which demonstrates the advantage of KG-CoI in reducing hallucinations.

\end{itemize}
\vspace{-0.07in}
\section{Related Work}

\subsubsection{Retrieval-augmented Generation.}
The integration of external knowledge into large language models (LLMs) has become an increasingly explored area of research, particularly for enhancing the accuracy and reliability of generated content \cite{gao2023retrieval,zhao2024retrieval}. This approach, known as retrieval-augmented generation (RAG), helps mitigate issues like hallucinations by grounding LLM outputs in relevant and accurate information from external sources \cite{lewis2020retrieval,guu2020retrieval,borgeaud2022improving,izacard2020leveraging}. In various domains, particularly biomedical and scientific research, retrieval-augmented methods have proven effective in improving LLM performance on tasks such as question answering and claim verification \cite{zakka2024almanac,xiong2024benchmarking,liu2024retrieval}.

Recent advancements have further refined these methods by incorporating knowledge graphs (KGs) into the retrieval process, addressing limitations such as the omission of rare but crucial information and the overrepresentation of frequently seen concepts. For instance, tools like KRAGEN utilize KGs to enhance LLM capabilities by structuring the retrieval process through graph-based reasoning \cite{matsumoto2024kragen}. Similarly, hybrid approaches that combine KG-based retrieval with traditional text embedding methods have shown promise in handling the long tail of biomedical knowledge, providing a more balanced and comprehensive retrieval of information \cite{delile2024graph}. In our work, we leverage the authoritative knowledge of existing domain-specific KGs for the generation of new scientific hypotheses, which 
demonstrates the potential of integrating structured knowledge with LLMs to inspire novel insights that might be valuable for scientific research.

\subsubsection{Chain-of-thought Based Reasoning.}
LLM reasoning can be powered by the Chain-of-Thoughts (CoT) prompting \cite{wei2022chain} technique, which involves prompting the LLM to generate intermediate reasoning steps when answering user questions. This technique helps the LLM generate much more accurate results than standard prompting. While the reasoning paths generated by CoT are a useful mechanism to explain why the LLM generated an answer, some studies show it is not faithful to the LLM's inner reasoning \cite{liu2023trustworthy}. Additionally, there can be reasoning steps that sound plausible but factually or logically flawed \cite{liu2023trustworthy}. Several studies demonstrate the effectiveness of incorporating external knowledge with CoT prompting to obtain more accurate, factually grounded answers while improving the explainability of reasoning paths \cite{luo2023reasoning, trivedi2022interleaving, wen2023mindmap}. Following CoT, advanced frameworks are further proposed to enhance LLM for accurate and reliable outputs \cite{wang2022self,yao2024tree,besta2024graph}, which leverage the internal knowledge of LLMs for complex reasoning.

\subsubsection{Hypothesis Generation.}
Hypothesis generation has been pursued as the task of mining meaningful implicit association between disjoint concepts in decades, where the goal is to predict connections between different concepts identified within scientific literature \cite{sebastian2017emerging}. This process typically involves uncovering meaningful relationships between separate concepts by systematically analyzing existing publications. Traditional methods in this domain often focus on identifying these connections from a static view of the literature.
While these approaches are effective and can be rigorously tested, they generally assume that all relevant concepts are pre-existing in the literature and simply need to be linked. This approach lacks the ability to account for the contextual factors that researchers consider important during the hypothesis formation process, and it does not fully engage with the creative and generative aspects of scientific thinking.

Recently, attention has shifted towards using LLMs to generate hypotheses. For example, SciMON \cite{wang2023scimon} proposed a framework that utilizes prior scientific literature to fine-tune LLMs for the generation of novel ideas. Additionally, in \cite{zhou2024hypothesis}, a method was introduced that uses prompts with LLMs to iteratively generate hypotheses based on provided examples. Our approach differs by utilizing both structured and unstructured domain-specific knowledge from scientific resources while leveraging the reasoning capabilities of LLMs to explore new insights from the current knowledge. Moreover, we focus on the hallucination detection in the generated content, providing an end-to-end system from context retrieval to hypothesis evaluation.

\vspace{-0.07in}
\section{Methodology} \label{sec:method}
\begin{figure*}[h!]
    \centering
    \includegraphics[width=0.85\linewidth]{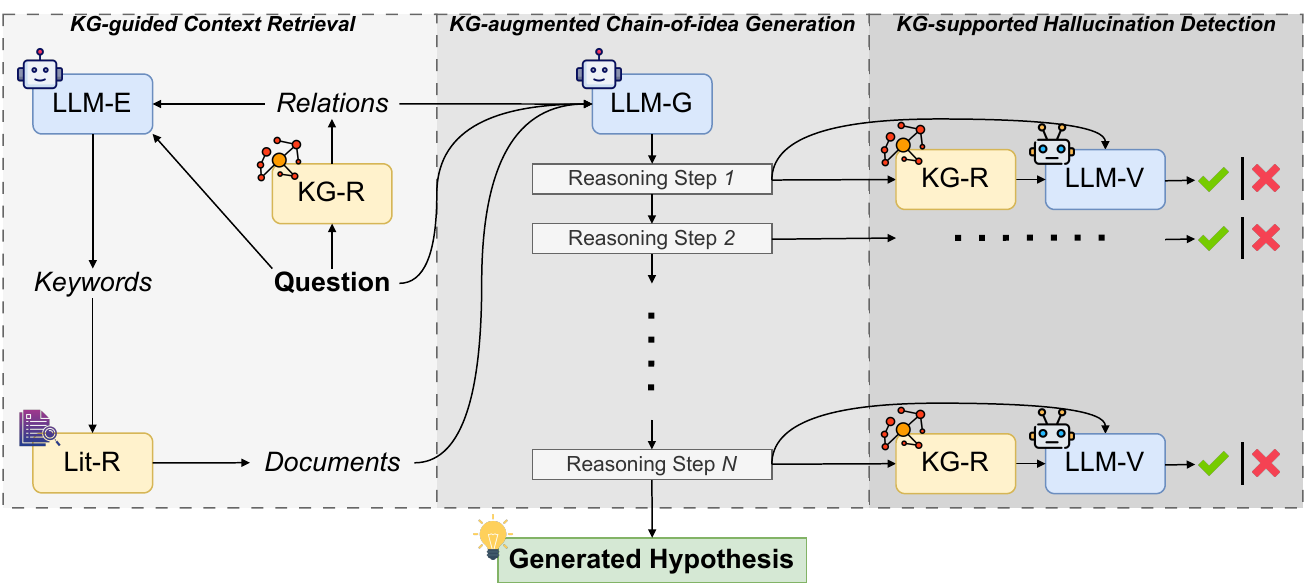}
    \caption{An overview of our proposed KG-CoI for knowledge-grounded hypothesis generation. ``KG-R'' and ``Lit-R'' are retrievers for scientific knowledge graphs (KGs) and literature, respectively. ``LLM-E'', ``LLM-G'', and ``LLM-V'' are LLM agents for query enrichment, hypothesis generation, and claim verification, respectively.}
    \label{fig:overview}
\end{figure*}

Figure \ref{fig:overview} shows the overview of our KG-CoI system. There are three different modules in KG-CoI, including the KG-guided context retrieval, the KG-augmented chain-of-idea generation, and the KG-supported hallucination detection. KG-CoI provides an end-to-end solution to augment the hypothesis generation of LLMs with KG information while providing an explicit analysis of hallucinations in the generated content. The details of each module are introduced in the remaining parts of this section.

\subsection{KG-guided Context Retrieval}

While retrieval-augmented generation \cite{lewis2020retrieval} provides an opportunity to enhance LLM output with external information from domain-specific corpora, the generation of scientific hypotheses is a challenging task where the relevant information of the new scientific knowledge can hardly be found in the existing literature. Thus, we propose to augment the hypothesis generation capability of LLMs using a KG-guided context retrieval, instead of the vanilla information retrieval from literature.

\subsubsection{Knowledge Enhancement with KG.}
The first key step of our KG-guided context retrieval is to enhance the given question using authoritative knowledge from KG. Specifically, for a given question \(\mathcal{Q}\) about the possible scientific facts between two entities \(e_h\) and \(e_t\), we can search for all \(k\)-step relation chains in a knowledge graph \(\mathcal{G}\) such that \(e_h\) and \(e_t\) are linked by
\begin{equation}
e_h \overset{r_1}{\longleftrightarrow} e_1 \overset{r_2}{\longleftrightarrow} e_2 \dots \overset{r_k}{\longleftrightarrow} e_t,
\end{equation}
where the relations on the bidirectional arrows connect two adjacent entities. The directions of the relations are defined by the KG selected. As shown in Figure \ref{fig:overview}, KG-CoI uses a KG-based retriever (KG-R) to search for neighbor relations for the given question, which is implemented with a breadth-first search strategy.
The relation chains will then be used as external knowledge from KG to augment the context retrieval for the original question. For simplicity, we will refer to the retrieved relations as \(\mathcal{R}\) in our later descriptions.

\subsubsection{Query Enrichment with KG.}
Given the input question and the retrieved relations from KG, we propose to enrich the user query by generating keywords using all existing information, which can facilitate further information retrieval from the scientific literature. Formally, given an input question \(\mathcal{Q}\) and the retrieved KG relations \(\mathcal{R}\), 
the keywords \(w_1,\cdots, w_T\) will be generated as
\begin{equation}
    w_1,\cdots, w_T  = \argmax_{w_1^*, \cdots, w_T^*} \mathbb{P}_{\texttt{LLM-E}}(w_1^*, \cdots, w_T^* | \mathcal{Q}, \mathcal{R}),
\end{equation}
where $\texttt{LLM-E}$ is an LLM agent that is required to generate search keywords to help answer the given question using existing information. In our KG-CoI, the query enrichment serves as an important step to naturally fuse the information of KG and the original question for context retrieval. We further test the importance of such a design, which is discussed in our ablation studies.

\subsubsection{Information Retrieval with Literature.}
After the query enrichment step, we perform information retrieval using the generated keywords to retrieve relevant documents from the scientific literature for the original question. For the biological domain explored in this paper, we use PubMed\footnote{https://pubmed.ncbi.nlm.nih.gov/} as the source of documents, including all biomedical abstracts in it. For the text retriever on PubMed, we select BM25 \cite{robertson2009probabilistic}, a sparse retriever that is based on lexicon comparisons. BM25 is a suitable choice for biological information retrieval, as domain-specific terms such as gene names may be wrongly tokenized by dense retrievers, leading to misunderstanding. We term a literature retrieval system with both a corpus and a retriever as ``Lit-R'', which takes the keywords \(w_1,\cdots, w_T\) as the input and outputs relevant documents \(\mathcal{D}\) from the given corpus.

\subsection{KG-augmented Chain-of-idea Generation}

We then prompt LLMs to perform a KG-augmented chain-of-idea generation, which is the core part of our KG-CoI system. The hypothesis generation of LLMs is augmented by KG from two different perspectives. First, we directly add the retrieved neighbor relations \(\mathcal{R}\) from KG to the input context of LLMs, making them knowledgeable of related information given by KG. Second, the retrieved documents from scientific literature are also provided to the LLMs for the augmentation of text generation. As described in the previous subsection, the retrieval of knowledge from literature is also guided by the relevant KG information, resulting in an implicit impact on the final output.

Additionally, we prompt LLMs to generate their predictions step by step as a chain of ideas (CoI). As discussed in the existing literature, prompting LLMs to have step-by-step thinking would help them dive deep into the given question and organize its answer in a logical way \cite{wei2022chain}. Moreover, it is crucial for our system to have a chain of ideas as the output of LLMs, which will be further used in our last module to analyze the hallucination and the confidence of LLMs in the generated contents.

Formally, given the original question \(\mathcal{Q}\), the retrieved KG relations \(\mathcal{R}\), and the retrieved documents \(\mathcal{D}\) from literature, a chain of ideas for the new scientific hypothesis can be generated by

\begin{equation} \label{eq:chain-of-ideas} \small
\left\{
\begin{aligned}
    s_1  &= \argmax_{s_1^*} \mathbb{P}_{\texttt{LLM-G}}(s_1^* | \mathcal{Q}, \mathcal{R}, \mathcal{D}), \\
    s_2  &= \argmax_{s_2^*} \mathbb{P}_{\texttt{LLM-G}}(s_2^* | \mathcal{Q}, \mathcal{R}, \mathcal{D}, s_1),\\
    & \cdots \\
    s_N  &= \argmax_{s_N^*} \mathbb{P}_{\texttt{LLM-G}}(s_N^* | \mathcal{Q}, \mathcal{R}, \mathcal{D}, s_1, \cdots, s_{N-1}),\\
    \mathcal{H}  &= \argmax_{\mathcal{H}^*} \mathbb{P}_{\texttt{LLM-G}}(\mathcal{H}^* | \mathcal{Q}, \mathcal{R}, \mathcal{D}, s_1, \cdots, s_{N}),
\end{aligned}
\right.
\end{equation}
where \(s_1,\cdots,s_N\) are the step-by-step ideas (or step-by-step claims) and \(\mathcal{H}\) is the final hypothesis generated given the chain of ideas. \(\texttt{LLM-G}\) is an LLM agent designed to generate new scientific hypotheses with the given information. 
While Formula \ref{eq:chain-of-ideas} describes a process of greedy search in the generation of ideas, we also explore the potential of running KG-CoI multiple times with randomness and examine its self-consistency in the answer prediction \cite{wang2022self}, which will be compared in detail in our Experiments.

\subsection{KG-supported Hallucination Detection}
While the KG-guided context retrieval and the KG-augmented chain-of-idea generation modules have already enhanced LLMs and provided the final prediction of the hypothesis, it is also important to explore the hallucinations in the generated content and examine how reliable the hypothesis is. With the output organized as a chain of ideas in our KG-CoI system, we propose to verify the correctness of each generated reasoning step using the information from a domain-specific KG. 

For each reasoning step \(s_i\) in the chain of ideas defined in Formula \ref{eq:chain-of-ideas}, we first identify all \(B\) biological entities \(e_{i1}, \cdots, e_{iB}\) in the claim using a named entity recognition (NER) tool tailored for domain-specific research. We then define the correctness of the claim \(s_i\) given the KG \(\mathcal{G}\) as a boolean variable with values 0 and 1. The correctness of \(s_i\) will be 1 if \(\exists j, k \in \{1,\cdots,B\}\) and the relation \( r_{ijk}\) such that
\begin{equation}
(e_{ij}, r_{ijk}, e_{ik}) \in \mathcal{G}
\end{equation}
and 
\begin{equation}
\texttt{LLM-V}((e_{ij}, r_{ijk}, e_{ik}), s_i) = 1.
\end{equation}
\(\texttt{LLM-V}\) is an LLM agent designed for claim verification given a triple of head and tail entities as well as their relation, which outputs 1 if the claim can be verified by the relation triple else 0.

In practice, KG-CoI uses the KG retriever ``KG-R'' defined in the KG-guided context retrieval phase, leveraging its abilities to find the direct links among entities that appear in the current claim. The retrieved relations will then be examined iteratively to check if any of them can support the claim made by LLMs in their chain of ideas. We only consider the direct link among entities because, in the ideal case, the reasoning steps of the hypothesis generation should be composed of simple scientific facts that can be verified by authoritative KG information. While the chain of ideas enables LLMs to have a deep analysis of the original question, the knowledge it uses for each reasoning step should be simple and easy to verify.

\begin{algorithm}[ht!]
\caption{The algorithm of KG-CoI for scientific hypothesis generation}
\textbf{Input}: A scientific question \(\mathcal{Q}\), Knowledge Graph \(\mathcal{G}\), Scientific Literature Corpus \(\mathcal{C}\), Maximum relation chain length \(k\)\\
\textbf{Output}: A scientific hypothesis \(\mathcal{H}\) with confidence score
\begin{algorithmic}[1]
\State \textbf{Step 1: KG-guided Context Retrieval}
\State Retrieve \(k\)-step relation chains \(\mathcal{R}\) from \(\mathcal{G}\) for entities in \(\mathcal{Q}\) using the KG retriever ``KG-R''
\State Generate enriched query keywords \(w_1,\cdots, w_T\) using \(\texttt{LLM-E}\) based on \(\mathcal{Q}\) and \(\mathcal{R}\)
\State Retrieve relevant documents \(\mathcal{D}\) from \(\mathcal{C}\) using a literature retriever ``Lit-R'' with keywords \(w_1,\cdots, w_T\)

\State \textbf{Step 2: KG-augmented Chain-of-idea Generation}
\State Generate step-by-step ideas \(s_1,\cdots,s_N\) for a new hypothesis \(\mathcal{H}\) using \(\texttt{LLM-G}\), incorporating \(\mathcal{Q}\), \(\mathcal{R}\), and \(\mathcal{D}\)
\State Formulate the final hypothesis \(\mathcal{H}\) based on the chain of ideas

\State \textbf{Step 3: KG-supported Hallucination Detection}
\For {each step \(s_i\) in \(\{s_1, \cdots, s_N\}\)}
    \State Identify entities \(e_{i1}, \cdots, e_{iB}\) in \(s_i\) using NER
    \State Check correctness of \(s_i\) by verifying relations \( (e_{ij}, r_{ijk}, e_{ik}) \in \mathcal{G}\) using \(\texttt{LLM-V}, \forall j,k \in \{1,\cdots,B\}\)
    \State Assign correctness score \( \text{correctness}(s_i) \) based on KG validation
\EndFor
\State Calculate confidence score using Formula \ref{eq:confidence}

\State \textbf{Return} the final hypothesis \(\mathcal{H}\) and its confidence score
\end{algorithmic}
\label{alg:algorithm}
\end{algorithm}

\begin{table*}[ht!]
    \centering
    \begin{tabular}{lcccccccccc}
    \toprule
    \multirow{3}{*}{\bf LLM} & \multirow{3}{*}{\bf Method} & \multirow{3}{*}{\bf Knowledge} & \multicolumn{3}{c}{\bf Greedy Search} & \multicolumn{3}{c}{\bf Self Consistency} \\
    \cmidrule(lr){4-6} \cmidrule(lr){7-9}
    & & & Accuracy & F1 & Confidence & Accuracy & F1 & Confidence \\
    \midrule
    Llama-3.1-8B & Direct & No & 47.00 & 48.71 & 00.00 & 65.00 & 70.86 & 00.00 \\
    Llama-3.1-8B & CoT & No & 56.67 & 56.61 & 40.22 & 56.67 &  
64.92 & \textbf{41.36} &  \\
    Llama-3.1-8B & RAG & Yes & 68.67 & 69.83 & 37.44 & 65.00 &  
70.86 & 38.35 &  \\
    Llama-3.1-8B & KG-CoI & Yes  & \bf 70.33 & \bf 70.42 & \textbf{43.46} & \textbf{66.67} & \textbf{72.63} & 40.46 &  \\
    \midrule
    Llama-3.1-70B & Direct & No & 72.00 & 71.94 & 00.00 & 71.67 & 72.00 & 00.00 \\
    Llama-3.1-70B & CoT & No & 73.67 & 73.77 &  \bf 35.23 & 71.33 & 71.58 & \textbf{34.97} &  \\
    Llama-3.1-70B & RAG & Yes & 73.33 & 73.50 & 35.02 & 72.33 & 73.12 & 34.58 &  \\
    Llama-3.1-70B & KG-CoI & Yes  & \bf 79.33 & \bf 79.52 & 30.78 & \textbf{81.00} & \textbf{81.92} & 26.24 &  \\
    \midrule
    GPT-4o-mini & Direct & No & 70.00 & 69.45 & 00.00 & 69.33 & 
 69.71 & 00.00 \\
    GPT-4o-mini & CoT & No & 73.00 & 72.61 & 39.28 & 73.33 & 73.59 & 39.21 &  \\
    GPT-4o-mini & RAG & Yes & 76.67 & 76.55 & 40.61 & 76.67 &  76.70 & 40.04  &  \\
    GPT-4o-mini & KG-CoI & Yes & \textbf{82.67} & \textbf{82.56} & \textbf{43.87} & \textbf{84.00} & \textbf{84.27} & \textbf{44.24} &  \\
    \midrule
    GPT-4o & Direct & No & 73.33 & 73.40 & 00.00 & 74.00 & 74.37 & 00.00 \\
    GPT-4o & CoT & No & 74.33 & 74.26 & 34.41 & 75.67 & 75.68 & 34.93 &  \\
    GPT-4o & RAG & Yes & 75.67 & 75.97 & 37.74 & 74.33 & 74.74 & 36.21 &  \\
    GPT-4o & KG-CoI & Yes & \textbf{86.00} & \textbf{85.83} & \bf 44.24 & \textbf{86.33} &  \textbf{86.17} & \textbf{41.66} &  \\
    \bottomrule
    \end{tabular}
    \caption{Comparison of our proposed KG-CoI system and baseline methods on hypothesis generation for biological knowledge. ``Knowledge'' denotes if the method is augmented with external biological knowledge. All scores are percentages.}
    \label{tab:main_table}
\end{table*}

After computing the correctness of each reasoning step in the chain of ideas, KG-CoI can summarize the overall confidence of a generated hypothesis as
\begin{equation} \label{eq:confidence}
    \text{confidence}(\mathcal{H}) = \frac{1}{N}\sum_{i=1}^N \text{correctness}(s_i).
\end{equation}

The measure of confidence can reflect the hallucinations in the generated content in terms of knowledge from a KG. If the confidence of a hypothesis is high, it means most reasoning steps in the chain of ideas can be verified by an external KG, indicating a high probability for the overall analysis to be reliable. If the confidence is low for a generated hypothesis, it shows most reasoning steps are unverifiable, calling for additional cautions when using the hypothesis.

The overall algorithm of our KG-CoI system is presented in Algorithm \ref{alg:algorithm}.
\section{Experiments}

\subsection{Experimental Settings}
To simulate the process of generating novel hypotheses, we use the knowledge graph (KG) of PubTator3 \cite{wei2024pubtator} and remove a set of relations from it to examine the capabilities of LLMs in hypothesizing the hidden relations using other existing knowledge in the KG. The constructed dataset mimics real-world scenarios where LLMs need to analyze existing knowledge and hypothesize new scientific facts, while providing a ground truth for model evaluation and comparison. The constructed hypothesis generation dataset contains 300 instances, where each of the three target classes (``stimulate'', ``inhibit'', ``no relation'') has 100 instances. More details about the dataset construction can be found in the Appendix.

\begin{table*}[h!]
    \centering
    \begin{tabular}{p{16.5cm}}
    \toprule
    \bf \makecell[c]{Hypothesis generation question} \\
    \midrule
    \textbf{Input}: Can we hypothesize a key relation between Carbon Tetrachloride and SLC17A5? \\
    \textbf{Ground Truth}: stimulate \\
    \midrule
    \rowcolor[RGB]{234, 238, 234}
    \bf \makecell[c]{Direct} \\
    \midrule
    \textbf{Answer}: no relation \textcolor{red}{\xmark} \\
    \midrule
    \rowcolor[RGB]{234, 238, 234}
    \bf \makecell[c]{CoT} \\
    \midrule
    \textbf{Reasoning}: Carbon Tetrachloride is a toxic compound that can cause cellular damage, and SLC17A5 is a gene involved in the transport of organic anions, including potentially toxic compounds. \sethlcolor{lightred}\hl{There is evidence that SLC17A5 can be involved in the transport of carbon tetrachloride, which can lead to cellular damage and toxicity.} \\
    \textbf{Answer}: inhibit \textcolor{red}{\xmark} \\
    \midrule
    \rowcolor[RGB]{234, 238, 234}
    \bf \makecell[c]{RAG} \\
    \midrule
    \textbf{Retrieved documents}: [...] \\
    \textbf{Reasoning}: Carbon Tetrachloride is a toxic compound that can cause damage to the liver and other organs. SLC17A5 is a gene that encodes for a protein called sialin, which is involved in the transport of sialic acid in the lysosome. \sethlcolor{lightred}\hl{There is no direct evidence that Carbon Tetrachloride interacts with SLC17A5.} \\
    \textbf{Answer}: no relation \textcolor{red}{\xmark} \\
    \midrule
    \rowcolor[RGB]{234, 238, 234}
    \bf \makecell[c]{KG-CoI (Ours)} \\
    \midrule
    \textbf{Neighbor relations}:     (Carbon Tetrachloride, cause, Liver Failure), (Liver Failure, associate, SLC17A5), [...] \\
    \textbf{Retrieved documents}: [...] \\
    \textbf{Reasoning}: \hl{Carbon Tetrachloride is a chemical that causes liver injury, and SLC17A5 is a gene associated with liver diseases, including liver failure and fibrosis.} The relationship between Carbon Tetrachloride and SLC17A5 is that Carbon Tetrachloride causes liver injury, which is associated with SLC17A5. \\
    \textbf{Answer}: stimulate \greencheck \\
    \bottomrule
    \end{tabular}
    \caption{An example of how KG-CoI helps Llama-3.1-8B generate correct hypotheses and reduce hallucinations. \hl{Verified reasoning steps} and \sethlcolor{lightred}\hl{misinformation} are highlighted. The correct answer is marked with ``\greencheck'' and the wrong answers are marked with ``\textcolor{red}{\xmark}''.}
    \label{tab:case_study}
\end{table*}

We choose direct prompting (Direct), chain-of-thought prompting (CoT; \citeauthor{wei2022chain}, \citeyear{wei2022chain}), and retrieval-augmented generation (RAG; \citeauthor{lewis2020retrieval}, \citeyear{lewis2020retrieval}) as the baselines for comparison, exploring if KG-CoI helps LLMs better find out the correct potential scientific facts given existing information. Direct and CoT examine if LLMs can make correct predictions based on their own parametric knowledge. RAG shows how well LLMs perform with external knowledge from scientific literature only. For each LLM and setting, we test its performance with both the greedy search and the self-consistency across five runs \cite{wang2022self}. 
More ablation studies about different components in KG-CoI are presented in the ``Ablation Studies'' section.

For our KG-CoI, we implement the KG retriever KG-R using the KG given by PubTator3 \cite{wei2024pubtator} along with its named entity recognition tool, which can identify the exact entity ID given the term description. 
We choose the ``en\_core\_sci\_sm" model from ScispaCy \cite{neumann2019scispacy} to extract biological named entities from a complete sentence.
The retriever and corpus selected for the literature-based information retrieval system Lit-R are BM25 \cite{robertson2009probabilistic} and PubMed, respectively, as mentioned in the Methodology. For LLM agents used in our system, we choose GPT-4o-mini as the LLM-V agent for claim verification, since the hallucination detection tool should be fixed for a fair comparison. More discussion of the choice of LLM-V will be presented in the Appendix.
For the implementation of LLM-E and LLM-G agents which are responsible for the generation of the chain of ideas and final hypotheses, we select both commercial and open-source models with various sizes, including Llama-3.1-8B, Llama-3.1-70B, GPT-4o-mini, and GPT-4o.

For LLMs with each setting, we compute the correctness of answers using accuracy and F1 scores. For settings that provide a chain of ideas for hypothesis generation (CoT, RAG, KG-CoI), we evaluate their hallucinations using the proposed KG-supported hallucinating detection tool in our KG-CoI system. The results of the hallucination detection will be summarized as ``Confidence'', indicating the proportion of claims verified by a given KG in an idea chain.
More details about our experimental settings can be found in the Appendix.

\subsection{Main Results}

Table \ref{tab:main_table} presents the main results of our experiments, showing how KG-CoI performs compared with other methods using different LLMs. We can observe from the table that KG-CoI consistently outperforms all other methods on different LLMs in terms of accuracy and F1. Specifically, the performance of LLMs on hypothesis generation gradually improve with the incorporation of reasoning capabilities (Direct \(\rightarrow\) CoT), knowledge from scientific literature (CoT \(\rightarrow\) RAG), and knowledge from KG (RAG \(\rightarrow\) KG-CoI). By comparing different LLMs, we can see that larger models (Llama-3.1-70B, GPT-4o) tend to perform better than smaller ones (Llama-3.1-8B, GPT-4o-mini) when using the same method for hypothesis generation. Interestingly, with the assistance of KG-CoI, the weakest LLM in our experiment (Llama-3.1-8B) can present an accuracy and F1 score close to the most advanced LLM (GPT-4o) in the ``Direct'' setting.

Moreover, Table \ref{tab:main_table} reveals that KG-CoI helps reduce hallucinations in LLM generation with more reasoning steps verified by domain-specific KG. While CoT examines the internal knowledge of LLMs, both RAG and KG-CoI augment the LLM hypothesis generation with external biological knowledge. It can be observed that RAG improves the confidence of hypotheses generation by GPT-4o-mini and GPT-4o, but does not show the same pattern on Llama-3.1. As the KG only contains authoritative and objective knowledge in the domain, the knowledge in literature may not have an exact match in KG. Thus, the retrieved documents from biological literature sometimes may not be verified by the KG, leading to the fluctuating confidence changes given by RAG in different LLMs. Nevertheless, with additional authoritative knowledge from KG, KG-CoI improves the model confidence on most LLMs examined, with a 3.30\% confidence increase on average compared to the CoT method.

\begin{table*}[h!]
    \centering
    \begin{tabular}{lcccccccccc}
    \toprule
    \multirow{3}{*}{\bf Setting} & \multicolumn{2}{c}{\bf Llama-3.1-8B} & \multicolumn{2}{c}{\bf Llama-3.1-70B} & \multicolumn{2}{c}{\bf GPT-4o-mini} & \multicolumn{2}{c}{\bf GPT-4o} \\
    \cmidrule(lr){2-3} \cmidrule(lr){4-5} \cmidrule(lr){6-7} \cmidrule(lr){8-9}
    & Accuracy & F1 & Accuracy & F1 & Accuracy & F1 & Accuracy & F1 \\
    \midrule    KG-CoI & \bf 70.33 & \bf 70.42 & \bf 79.33 & \bf 79.52 & \bf 82.67 & \bf 82.56 & \bf 86.00 & \bf 85.83 \\
    \textcolor{red}{\xmark} KG information & 65.00 & 65.66 & 73.33 & 73.24 & 74.33 & 74.29 & 74.67 & 74.93 \\
    \textcolor{red}{\xmark} Literature information & 60.67 & 59.96 & 75.33 & 75.34 & 76.00 & 75.53 & 83.33 & 83.35  \\
    \textcolor{red}{\xmark} Query enrichment & 64.67 & 65.01 & 79.00 & 79.18 & 78.67 & 78.51 & 83.00 & 83.19 \\
    \textcolor{red}{\xmark} Chain of thoughts & 62.00 & 62.19 & 77.00 & 76.99 & 71.67 & 72.39 & 76.00 & 75.73  \\
    \bottomrule
    \end{tabular}
    \caption{Ablation studies of different components in KG-CoI on various LLMs. ``KG-CoI'' denotes the full version of our proposed system. ``\textcolor{red}{\xmark}'' means the removal of specific components in KG-CoI.}
    \label{tab:ablation}
\end{table*}

In addition to the results for the greedy search of LLMs on the constructed dataset, we also examine if LLMs benefit from multiple runs on each instance using self-consistency \cite{wang2022self}. While Llama-3.1-8B presents an increased F1 score but a decreased accuracy with multiple runs, self-consistency is shown to be effective for KG-CoI on all other three LLMs in terms of accuracy and F1. On the measurement of hallucinations, it is shown that the comparison of methods in the self-consistency setting presents the same patterns as in the greedy search setting, reflecting the effectiveness of KG-CoI in reducing hallucinations. The results also reveal that compared to greedy search, self-consistency does not necessarily improve the confidence of generated hypotheses. This may be a result of output uncertainty caused by the introduced randomness in the multiple trials of the self-consistency setting as opposed to the consistent and deterministic nature of greedy search.

\subsection{Case Studies}

To understand how KG-CoI helps LLMs generate the correct hypothesis, we perform case studies on actual instances in our dataset to analyze the effect of components in KG-CoI on the final prediction. Table \ref{tab:case_study} shows an example where KG-CoI helps Llama-3.1-8B find the true relation between Carbon Tetrachloride and SLC17A5. While the direct prompting of the LLM provides a wrong answer, the CoT prompting results in an incorrect answer with hallucinated reasoning steps. As can be observed from the case, LLMs with CoT may hallucinate scientific knowledge that is not verified by existing KG. In contrast, RAG may fail to retrieve useful information from the literature to augment the hypothesis generation, leading to false negatives of the prediction.

By providing LLMs with neighbor information from the domain-specific KG, KG-CoI enables LLMs to reason on objective structured knowledge that may not be explicitly stated in the scientific literature. Table \ref{tab:case_study} shows the addition of KG knowledge helps Llama-3.1-8B build the reasonable logic chain to link different entities and find out the ground truth relation. Also, the use of both literature and KG information in KG-CoI provides verified knowledge from reliable sources, reducing the hallucinations in the generated content. 

\subsection{Ablation Studies}

As described in the Methodology, our KG-CoI is a multi-step hypothesis generation system with external knowledge from different sources. To illustrate how each component contributes to the entire system, we perform ablation studies to see how different settings affect the performance of the system. 
We first ablate the source of biological knowledge in KG-CoI. As both KG and literature information are used in our system, we test if the removal of any of them will lead to a performance drop. Specifically, the removal of KG information is performed by discarding the neighbor KG relations in both the query enrichment and the augmented generation steps. The literature information is removed by using the neighbor relations only to augment the hypothesis generation. We further examine a special setting with the removal of query enrichment, retrieving useful documents from literature based on the original question only instead of using keywords generated given both the question and the neighbor relations. The last setting in our ablation studies involves the removal of the chain of thoughts, testing how the system performs without prompting LLMs to think step-by-step.

The results of our ablation studies are presented in Table \ref{tab:ablation}. In general, the performance of KG-CoI drops with the removal of any component in our ablation studies, which is consistently observed on all LLMs. The importance of different components to the model performance is shown to be diverse in various LLMs. For example, the removal of KG information brings the most dramatic performance decrease on Llama-3.1-70B and GPT-4o, while it is the least important one with the minimal performance change on Llama-3.1-8B. We also discover from Table \ref{tab:ablation} that, from the LLM with the lowest accuracy (Llama-3.1-8B, 70.33\%) to the highest accuracy (GPT-4o, 86.00\%), the relative importance of the literature information changes from the most important to the least one on corresponding LLMs. While the removal of the literature information causes a performance drop of 9.66\% in Llama-3.1-8B, it only decreases the performance of GPT-4o by 2.67\%. Such a result can be interpreted by the fact that more advanced LLMs tend to have read more literature during training, leading to fewer needs for external literature information. The ablation studies demonstrate the necessity of various components in our KG-CoI system, which also provide additional insights into the LLMs used in our experiments.

Additional experimental results can be found in the Appendix.
\section{Conclusion}

We propose KG-CoI, a systematic approach to enhancing the scientific hypothesis generation capability of LLMs with domain-specific knowledge graphs (KGs), which include KG-guided context retrieval, KG-augmented chain-of-idea generation, and KG-supported hallucination detection.
Using a newly constructed hypothesis generation dataset introduced in this work, we demonstrate the effectiveness of KG-CoI in generating correct hypotheses and reducing hallucinations. Our ablation studies and case studies further justify the component designs in our system and illustrate how its predictions will be generated and used in real-world applications. This work paves the potential for researchers to utilize LLMs as a tool to verify results and generate reliable insights for future research.

\clearpage

\bibliography{aaai24}

\clearpage
\appendix
\section{Appendix}

\subsection{Data and Code Availability}
Our data and source code are available at \url{https://anonymous.4open.science/r/KG-CoI-C203/}.

\subsection{Details of Dataset Construction}

We created a high-quality test set focusing on achieving a balance between creating an environment that mirrors a real-world hypothesis generation task and ensuring effective quantitative evaluation. Each question within the evaluation dataset is constructed from a subgraph extracted from PubTator3 \cite{wei2024pubtator}, containing information supported by a repository of biomedical literature. 

In PubTator3, a selected edge between two key nodes was deliberately removed to create a scenario of incomplete information. The model was then asked to hypothesize the relation between the two nodes. This approach accomplishes two objectives: it mimics the process of hypothesizing with incomplete information while expanding upon the current body of knowledge, and it provides a ground truth for effectively measuring the model's performance.

To select edges for our dataset, we focus on two relation types: ``stimulate" and ``inhibit". These terms are used in place of the labels given in PubTator3, ``positive\_correlate" and ``negative\_correlate," to ensure clarity in their definitions and reduce the risk of confusion in model predictions. We began by selecting a random initial node from within PubTator3. From this node, we identified a connected second node through one of its observed relations. Next, we analyzed the relations of this second node to find a third connected node, and continued to traverse the graph in this manner. To select each subsequent node, we sorted the relations of the previous node by the number of publications that observe each relation as a measure of relevance and significance, descending through the most frequently cited relations until a ``stimulate" or ``inhibit" is found. 

Once a relation is selected, we verified whether the opposing relation (i.e., if ``stimulate" is chosen, we verify ``inhibit", and vice versa) has a similar number of publications. Specifically, for a selection to be valid, the opposite relation must have less than half the number of publications as the selected relation. This process was repeated until 100 samples of stimulate or inhibit relations had been identified. To find ``no\_relation" pairings, we randomly selected nodes from those involved in previously identified ``stimulate" or ``inhibit" relations and verified the absence of a direct connection between them, continuing this process until 100 ``no\_relation" samples were obtained.

\subsection{Discussion on Hallucination Detection}

In the implementation of our KG-CoI system, we select GPT-4o-mini as the model for the \texttt{LLM-V} agent to verify if an LLM-generated claim can be supported by an existing relation triple in a domain-specific knowledge graph.

To justify the choice of GPT-4o-mini in the current implementation, we randomly sample 100 instances of (claim, relation triple) from the actual hallucination detection process, and manually annotate if each instance contains a supported claim. In addition to the results from GPT-4o-mini, we also test GPT-3.5 and GPT-4o to see how GPT-4o-mini performs compared with them.

Table \ref{tab:annotation_exact_match} presents the exact match ratio of different annotators. Among the three examined LLMs, GPT-4o aligns the best with human annotators while GPT-3.5-turbo performs the worst. However, considering the prices of the LLMs, GPT-4o-mini turns out to be the most cost-effective choice with a low price but a good performance. Thus, we select GPT-4o-mini to be the \texttt{LLM-V} in our experiments for hallucination detection.

\begin{table}[h!]
    \centering
    \begin{tabular}{l|ccccc}
    \toprule
         & H1 & H2 & 3.5 & 4o-mini & 4o \\
        \midrule
        H1 & 1.00 & 0.90 & 0.70 & 0.74 & 0.76 \\
        H2 & 0.90 & 1.00 & 0.66 & 0.72 & 0.82 \\
        3.5 & 0.70 & 0.66 & 1.00 & 0.82 & 0.74 \\
        4o-mini & 0.74 & 0.72 & 0.82 & 1.00 & 0.80 \\
        4o & 0.76 & 0.82 & 0.74 & 0.80 & 1.00 \\
    \midrule
    Price & -- & -- & \$0.50 & \$0.15 & \$5.00 \\
    \bottomrule
    \end{tabular}
    \caption{Exact match of different annotations on claim verification. ``H1'' stands for the first human annotator. ``H2'' stands for the second human annotator. ``3.5'', ``4o-mini'', ``4o'' denote GPT-3.5-Turbo, GPT-4o-mini, and GPT-4o, respectively. The prices for 1M input tokens are listed for the tested LLMs.}
    \label{tab:annotation_exact_match}
\end{table}

\subsection{Additional Results on Self-consistency Scaling}

In Table \ref{tab:main_table} of the main paper, we demonstrate that the self-consistency \cite{wang2022self} of multiple runs helps KG-CoI find out more accurate hypotheses with increased accuracy and F1 scores. To explore if such an improvement grows with the scaling number of runs in self-consistency, we perform further experiments on KG-CoI to evaluate its performance with the number of runs \(N\) in self-consistency to be \(N=1,5,10,15\).

\begin{figure*}
    \centering
    \includegraphics[width=0.68\linewidth]{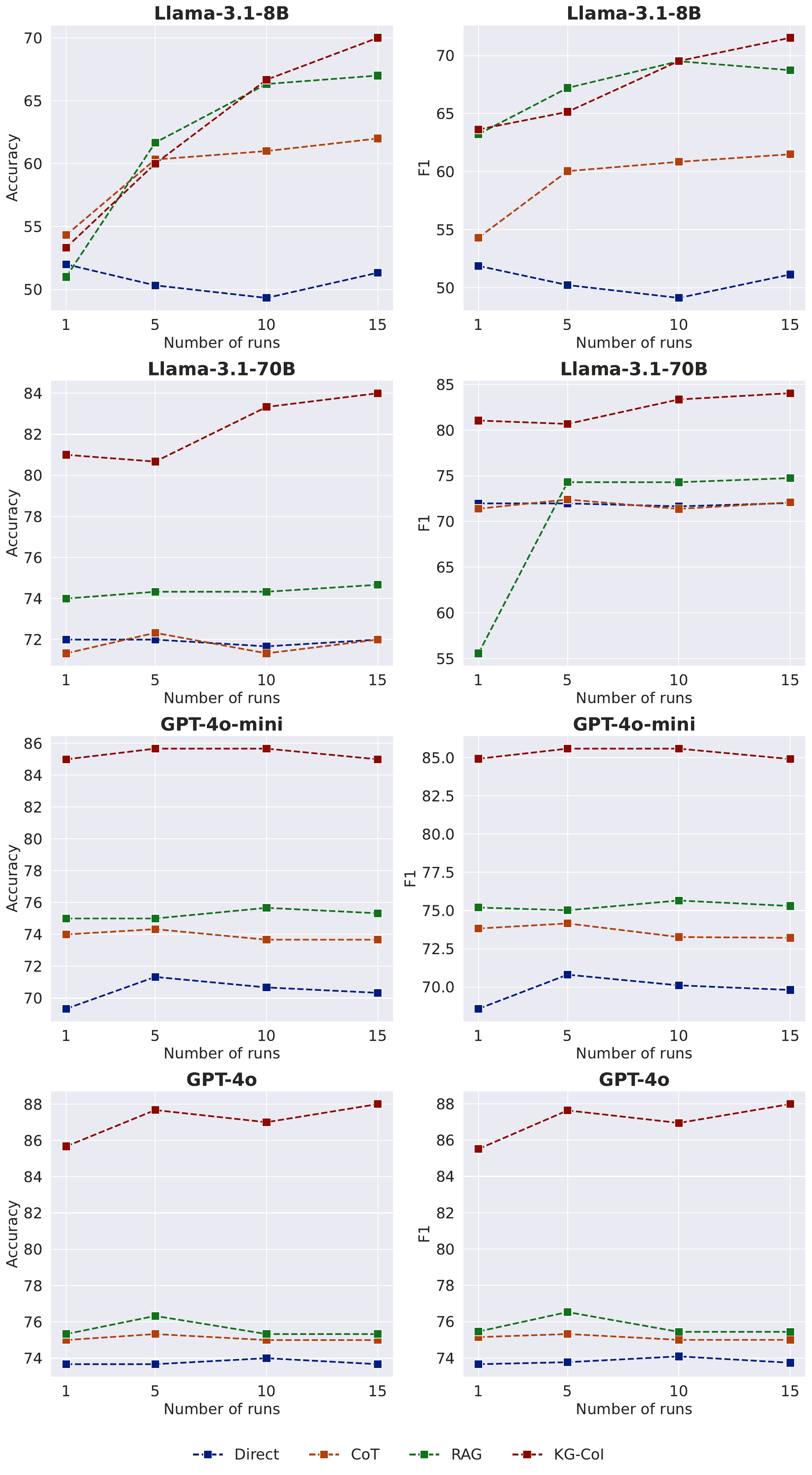}
    \caption{Performance of various methods with different numbers of runs used in the self-consistency setting.}
    \label{fig:sc_scaling}
\end{figure*}

Figure \ref{fig:sc_scaling} shows the 
performance of various LLMs and methods when we increase the number of runs in the self-consistency setting. From the results, we can observe that KG-CoI consistently outperforms other compared methods when we scale up the runs in self-consistency. Moreover, the LLM performance tends to improve with the increased number of runs, especially in weak models such as Llama-3.1-8B and Llama-3.1-70B. These results demonstrate the potential of KG-CoI to be further improved by increasing the number of runs when using the self-consistency technique.

\subsection{Prompt Templates in Experiments}

To improve the reproducibility of our work and facilitate the follow-up research based on KG-CoI, we provide all prompt templates we used in our experiments in this section, which are shown in Figures \ref{fig:prompt_direct} - \ref{fig:prompt_llm_v}. For all tested methods, we add a simple example question in the prompts to instruct LLMs to perform this new hypothesis generation task, which was manually designed by human annotators.

\begin{figure*}[h]
\begin{AIbox}{Prompt template for Direct prompting of LLMs}
You are a leading scientist tasked with hypothesizing interactions between two given biochemical entities in the format ``Answer: [`relation']'' without the quotes. The answer choices are [`inhibit'], [`stimulate'], or [`no\_relation'].\\

Example: 

Question: Can we hypothesize a key relation between \@GENE\_JAK1 and \@CHEMICAL\_ruxolitinib?\\Output your answer in the format ``Answer: [your\_answer (among `inhibit', `stimulate', and `no\_relation')]".\\
Answer: [`inhibit']\\

Question: \verb|{{question}}|

Output your answer in the format ``Answer: [your\_answer (among `inhibit', `stimulate', and `no\_relation')]".\\
\end{AIbox}
\caption{Prompt template for Direct prompting of LLMs.}
\label{fig:prompt_direct}
\end{figure*}

\begin{figure*}[h]
\begin{AIbox}{Prompt template for CoT prompting of LLMs}
You are a leading scientist tasked with hypothesizing interactions between two given biochemical entities in the format Reasoning:, then Answer:. You must strictly follow this structure. Each reasoning step should be verifiable using a knowledge graph. The answer choices are inhibit, stimulate, or no\_relation.\\

Example: 

Question: Can we hypothesize a key relation between \@GENE\_JAK1 and \@CHEMICAL\_ruxolitinib?

Output your answer in the format ``Reasoning: your\_reasoning\_steps", then ``Answer: [your\_answer (among `inhibit', `stimulate', and `no\_relation')]".\\Reasoning: Ruxolitinib is a JAK1 inhibitor. Additionally, Ruxolitinib has been shown to treat diseases associated with JAK1 mutations.

Answer: [`inhibit']\\

Question: \verb|{{question}}|

Output your answer in the format ``Reasoning: your\_reasoning\_steps", then ``Answer: [`inhibit'],", ``Answer: [`stimulate']", or ``Answer: [`no\_relation']".
\end{AIbox}
\caption{Prompt template for CoT prompting of LLMs.}
\label{fig:prompt_cot}
\end{figure*}

\begin{figure*}[h]
\begin{AIbox}{Prompt template for RAG prompting of LLMs}
You are a leading scientist tasked with hypothesizing interactions between two given biochemical entities in the format Reasoning:, then Answer:. You must strictly follow this structure. Each reasoning step should be verifiable using a knowledge graph. The answer choices are inhibit, stimulate, or no\_relation.\\

Example: 

Context: {context} Question: Can we hypothesize a key relation between \@GENE\_JAK1 and \@CHEMICAL\_ruxolitinib?\\Output your answer in the format ``Reasoning: your\_reasoning\_steps", then ``Answer: [your\_answer (among `inhibit', `stimulate', and `no\_relation')]".

Reasoning: Ruxolitinib is a JAK1 inhibitor. Additionally, Ruxolitinib has been shown to treat diseases associated with JAK1 mutations.

Answer: [`inhibit']\\

Question: \verb|{{question}}|

Output your answer in the format ``Reasoning: your\_reasoning\_steps", then ``Answer: [`inhibit'],", ``Answer: [`stimulate']", or ``Answer: [`no\_relation']". Make sure to include the brackets.
\end{AIbox}
\caption{Prompt template for RAG prompting of LLMs.}
\label{fig:prompt_rag}
\end{figure*}

\begin{figure*}[h]
\begin{AIbox}{Prompt template for \texttt{LLM-E} in KG-CoI}
You are an expert tasked with constructing a query to find documents that will answer a given question. Look it up as if you are creating a google search. Your output must only contain the keywords, nothing else.\\

You are an expert tasked with constructing a query to find documents that will answer a given question. Look it up as if you are creating a google search. Your output must only contain the keywords, nothing else, so do not say ``Here are the relevant keywords: " or anything of that nature. Additionally, you are given several relations that may assist in creating the query. Identify which connections from the list are the strongest and use them to construct the query. Your output will be fed directly into the retriever, so ensure it is in natural language format.\\

Context: \verb|{{context}}|

Question: \verb|{{question}}|
\end{AIbox}
\caption{Prompt template for  \texttt{LLM-E} in KG-CoI.}
\label{fig:prompt_llm_e}
\end{figure*}

\begin{figure*}[h]
\begin{AIbox}{Prompt template for \texttt{LLM-A} in KG-CoI}
You are a leading scientist tasked with hypothesizing interactions between two given biochemical entities in the format Reasoning:, then Answer:. You must strictly follow this structure. Each reasoning step should be verifiable using a knowledge graph. The answer choices are inhibit, stimulate, or no\_relation.\\

Example: 

Context: {context} Question: Can we hypothesize a key relation between \@GENE\_JAK1 and \@CHEMICAL\_ruxolitinib?\\Output your answer in the format ``Reasoning: your\_reasoning\_steps", then ``Answer: [your\_answer (among `inhibit', `stimulate', and `no\_relation')]".

Reasoning: Ruxolitinib is a JAK1 inhibitor. Additionally, Ruxolitinib has been shown to treat diseases associated with JAK1 mutations.

Answer: [`inhibit']\\

Question: \verb|{{question}}|

Output your answer in the format ``Reasoning: your\_reasoning\_steps", then ``Answer: [`inhibit'],", ``Answer: [`stimulate']", or ``Answer: [`no\_relation']". Make sure to include the brackets.
\end{AIbox}
\caption{Prompt template for  \texttt{LLM-A} in KG-CoI.}
\label{fig:prompt_llm_a}
\end{figure*}

\begin{figure*}[h]
\begin{AIbox}{Prompt template for \texttt{LLM-V} in KG-CoI}
You are tasked with finding if a relation extracted from a knowledge graph supports a given statement. For example, stating that a chemical has a negative correlation with a gene supports the chemical being an inhibitor, but stating that a chemical is simply 'associated' with the gene does not. Think it through, and use the format Reasoning: ... then Answer: [`...'] to structure your response. Possible answers include [`yes'] or [`no']\\

Relation: \verb|{{relation}}|

Statement: \verb|{{sentence}}|
\end{AIbox}
\caption{Prompt template for  \texttt{LLM-V} in KG-CoI.}
\label{fig:prompt_llm_v}
\end{figure*}

\subsection{Additional Case Studies}

In addition to the case study of Llama-3.1-8B shown in Table \ref{tab:case_study} of the main paper, we perform more case studies of other LLMs on the same input, which are presented in Tables \ref{tab:case_study_llama_70b}, \ref{tab:case_study_gpt_4o_mini}, \ref{tab:case_study_gpt_4o}. Similar to the study of Llama-3.1-8B, the results show that other LLMs also benefit from the additional KG information provided by KG-CoI, helping them hypothesize the correct relation between Carbon Tetrachloride and SLC17A5 that is not found by any other methods compared.

\begin{table*}[h!]
    \centering
    \begin{tabular}{p{16.5cm}}
    \toprule
    \bf \makecell[c]{Hypothesis generation question} \\
    \midrule
    \textbf{Input}: Can we hypothesize a key relation between Carbon Tetrachloride and SLC17A5? \\
    \textbf{Ground Truth}: stimulate \\
    \midrule
    \rowcolor[RGB]{234, 238, 234}
    \bf \makecell[c]{Direct} \\
    \midrule
    \textbf{Answer}: no relation \textcolor{red}{\xmark} \\
    \midrule
    \rowcolor[RGB]{234, 238, 234}
    \bf \makecell[c]{CoT} \\
    \midrule
    \textbf{Reasoning}: Carbon tetrachloride (CCl4) is a toxic compound that can cause liver damage. SLC17A5 is a gene that encodes a transporter involved in the transport of organic anions, including bile acids. Research has shown that CCl4 can alter the expression of SLC17A5 in the liver, leading to changes in bile acid transport. However, \sethlcolor{lightred}\hl{there is no direct evidence that CCl4 directly inhibits or stimulates SLC17A5}. Instead, the interaction is likely indirect, through CCl4-induced liver damage and subsequent changes in gene expression.
 \\
    \textbf{Answer}: no relation \textcolor{red}{\xmark} \\
    \midrule
    \rowcolor[RGB]{234, 238, 234}
    \bf \makecell[c]{RAG} \\
    \midrule
    \textbf{Retrieved documents}: [...] \\
    \textbf{Reasoning}: There is no known interaction between Carbon Tetrachloride and SLC17A5. Carbon Tetrachloride is a chemical compound that can cause liver damage, while SLC17A5 is a gene that encodes a protein involved in the transport of sialic acid. \sethlcolor{lightred}\hl{There is no established link between the two.} \\
    \textbf{Answer}: no relation \textcolor{red}{\xmark} \\
    \midrule
    \rowcolor[RGB]{234, 238, 234}
    \bf \makecell[c]{KG-CoI (Ours)} \\
    \midrule
    \textbf{Neighbor relations}:     (Carbon Tetrachloride, cause, Liver Failure), (Liver Failure, associate, SLC17A5), [...] \\
    \textbf{Retrieved documents}: [...] \\
    \textbf{Reasoning}: ((@CHEMICAL\_Carbon\_Tetrachloride has the relation cause with @DISEASE\_Liver\_Failure), which has the relation associate with @GENE\_SLC17A5). | ((@CHEMICAL\_Carbon\_Tetrachloride has the relation cause with @DISEASE\_Chemical\_and\_Drug\_Induced\_Liver\_Injury), which has the relation associate with @GENE\_SLC17A5). | ((@CHEMICAL\_Carbon\_Tetrachloride has the relation negative\_correlate with @CHEMICAL\_Silymarin), which has the relation negative\_correlate with @GENE\_SLC17A5). | ((@GENE\_SLC17A5 has the relation associate with @DISEASE\_Chemical\_and\_Drug\_Induced\_Liver\_Injury), which has the relation cause with @CHEMICAL\_Carbon\_Tetrachloride). | ((@GENE\_SLC17A5 has the relation associate with @DISEASE\_Liver\_Failure), which has the relation cause with @CHEMICAL\_Carbon\_Tetrachloride). | ((@GENE\_SLC17A5 has the relation associate with @DISEASE\_Fibrosis), which has the relation cause with @CHEMICAL\_Carbon\_Tetrachloride). | ((@GENE\_SLC17A5 has the relation associate with @DISEASE\_Liver\_Cirrhosis), which has the relation cause with @CHEMICAL\_Carbon\_Tetrachloride). | ((@GENE\_SLC17A5 has the relation stimulate with @DISEASE\_Chemical\_and\_Drug\_Induced\_Liver\_Injury), which has the relation cause with @CHEMICAL\_Carbon\_Tetrachloride). | ((@GENE\_SLC17A5 has the relation associate with @DISEASE\_Liver\_Diseases), which has the relation cause with @CHEMICAL\_Carbon\_Tetrachloride). | ((@GENE\_SLC17A5 has the relation stimulate with @DISEASE\_Liver\_Failure), which has the relation cause with @CHEMICAL\_Carbon\_Tetrachloride). | ((@GENE\_SLC17A5 has the relation associate with @DISEASE\_Fatty\_Liver), which has the relation cause with @CHEMICAL\_Carbon\_Tetrachloride). | ((@GENE\_SLC17A5 has the relation negative\_correlate with @CHEMICAL\_Silymarin), which has the relation negative\_correlate with @CHEMICAL\_Carbon\_Tetrachloride). | \hl{((@GENE\_SLC17A5 has the relation stimulate with @DISEASE\_Liver\_Diseases), which has the relation cause with @CHEMICAL\_Carbon\_Tetrachloride).} \\
    \textbf{Answer}: stimulate \greencheck \\
    \bottomrule
    \end{tabular}
    \caption{An example of how KG-CoI helps Llama-3.1-70B generate correct hypotheses and reduce hallucinations. \hl{Verified reasoning steps} and \sethlcolor{lightred}\hl{misinformation} are highlighted. The correct answer is marked with ``\greencheck'' and the wrong answers are marked with ``\textcolor{red}{\xmark}''.}
    \label{tab:case_study_llama_70b}
\end{table*}

\begin{table*}[!h]
    \centering
    \begin{tabular}{p{16.5cm}}
    \toprule
    \bf \makecell[c]{Hypothesis generation question} \\
    \midrule
    \textbf{Input}: Can we hypothesize a key relation between Carbon Tetrachloride and SLC17A5? \\
    \textbf{Ground Truth}: stimulate \\
    \midrule
    \rowcolor[RGB]{234, 238, 234}
    \bf \makecell[c]{Direct} \\
    \midrule
    \textbf{Answer}: no relation \textcolor{red}{\xmark} \\
    \midrule
    \rowcolor[RGB]{234, 238, 234}
    \bf \makecell[c]{CoT} \\
    \midrule
    \textbf{Reasoning}: SLC17A5 is a transporter protein involved in the transport of certain organic anions and is linked to the metabolism of various substances. Carbon Tetrachloride is a toxic compound that can cause liver damage and disrupt normal metabolic processes. \sethlcolor{lightred}\hl{There is no direct evidence or established interaction between} Carbon Tetrachloride and SLC17A5 that suggests a stimulating or inhibiting effect. Therefore, the relationship is likely to be non-existent.
 \\
    \textbf{Answer}: no relation \textcolor{red}{\xmark} \\
    \midrule
    \rowcolor[RGB]{234, 238, 234}
    \bf \makecell[c]{RAG} \\
    \midrule
    \textbf{Retrieved documents}: [...] \\
    \textbf{Reasoning}: Carbon Tetrachloride is known to cause liver damage and can lead to the impairment of various metabolic processes. However, \sethlcolor{lightred}\hl{there is no direct evidence or established interaction between Carbon Tetrachloride and the SLC17A5 gene}, which is primarily involved in sialic acid transport in lysosomes. Therefore, any hypothesized relation between Carbon Tetrachloride and SLC17A5 would not be supported by existing biochemical knowledge. \\
    \textbf{Answer}: no relation \textcolor{red}{\xmark} \\
    \midrule
    \rowcolor[RGB]{234, 238, 234}
    \bf \makecell[c]{KG-CoI (Ours)} \\
    \midrule
    \textbf{Neighbor relations}:     (Carbon Tetrachloride, cause, Liver Failure), (Liver Failure, associate, SLC17A5), [...] \\
    \textbf{Retrieved documents}: [...] \\
    \textbf{Reasoning}: \hl{Carbon Tetrachloride is known to cause liver damage and is associated with various liver diseases.} The gene SLC17A5 has been shown to have a relationship with liver diseases, including those induced by chemical exposure. Specifically, SLC17A5 is associated with conditions like Chemical and Drug-Induced Liver Injury, which is caused by Carbon Tetrachloride. Therefore, SLC17A5 is likely to be stimulated in response to the liver injury caused by Carbon Tetrachloride. \\
    \textbf{Answer}: stimulate \greencheck \\
    \bottomrule
    \end{tabular}
    \caption{An example of how KG-CoI helps GPT-4o-mini generate correct hypotheses and reduce hallucinations. \hl{Verified reasoning steps} and \sethlcolor{lightred}\hl{misinformation} are highlighted. The correct answer is marked with ``\greencheck'' and the wrong answers are marked with ``\textcolor{red}{\xmark}''.}
    \label{tab:case_study_gpt_4o_mini}
\end{table*}

\begin{table*}[h!]
    \centering
    \begin{tabular}{p{16.5cm}}
    \toprule
    \bf \makecell[c]{Hypothesis generation question} \\
    \midrule
    \textbf{Input}: Can we hypothesize a key relation between Carbon Tetrachloride and SLC17A5? \\
    \textbf{Ground Truth}: stimulate \\
    \midrule
    \rowcolor[RGB]{234, 238, 234}
    \bf \makecell[c]{Direct} \\
    \midrule
    \textbf{Answer}: no relation \textcolor{red}{\xmark} \\
    \midrule
    \rowcolor[RGB]{234, 238, 234}
    \bf \makecell[c]{CoT} \\
    \midrule
    \textbf{Reasoning}: Carbon Tetrachloride is a hepatotoxic chemical known to cause liver damage. SLC17A5 is a gene encoding a lysosomal sialic acid transporter, primarily involved in cellular transport mechanisms. \sethlcolor{lightred}\hl{There is no direct evidence or established pathway indicating that Carbon Tetrachloride interacts with or affects the function of SLC17A5.} \\
    \textbf{Answer}: no relation \textcolor{red}{\xmark} \\
    \midrule
    \rowcolor[RGB]{234, 238, 234}
    \bf \makecell[c]{RAG} \\
    \midrule
    \textbf{Retrieved documents}: [...] \\
    \textbf{Reasoning}: Carbon Tetrachloride is a chemical compound known for its hepatotoxic effects and is commonly used in research to induce liver damage. SLC17A5 is a gene encoding the sialin protein, which is involved in lysosomal transport of sialic acid. \sethlcolor{lightred}\hl{There is no known direct interaction or pathway linking Carbon Tetrachloride to the function or expression of SLC17A5.} \\
    \textbf{Answer}: no relation \textcolor{red}{\xmark} \\
    \midrule
    \rowcolor[RGB]{234, 238, 234}
    \bf \makecell[c]{KG-CoI (Ours)} \\
    \midrule
    \textbf{Neighbor relations}:     (Carbon Tetrachloride, cause, Liver Failure), (Liver Failure, associate, SLC17A5), [...] \\
    \textbf{Retrieved documents}: [...] \\
    \textbf{Reasoning}: The knowledge graph indicates that Carbon Tetrachloride causes various liver diseases, including liver failure, chemical and drug-induced liver injury, fibrosis, liver cirrhosis, and fatty liver. SLC17A5 is associated with these diseases and is shown to stimulate chemical and drug-induced liver injury, liver failure, and liver diseases. Additionally, SLC17A5 has a negative correlation with Silymarin, which in turn has a negative correlation with Carbon Tetrachloride. \hl{This suggests that SLC17A5 is involved in the pathways affected by Carbon Tetrachloride, and its activity is likely stimulated by the presence of Carbon Tetrachloride.} \\
    \textbf{Answer}: stimulate \greencheck \\
    \bottomrule
    \end{tabular}
    \caption{An example of how KG-CoI helps GPT-4o generate correct hypotheses and reduce hallucinations. \hl{Verified reasoning steps} and \sethlcolor{lightred}\hl{misinformation} are highlighted. The correct answer is marked with ``\greencheck'' and the wrong answers are marked with ``\textcolor{red}{\xmark}''.}
    \label{tab:case_study_gpt_4o}
\end{table*}

\end{document}